\documentclass{article}
\pdfoutput=1
\usepackage{spconf,amsmath,graphicx}
\usepackage{amsmath,amssymb}
\DeclareMathOperator{\E}{\mathbb{E}}
\usepackage{hyperref}

\title{Boundary of Distribution Support Generator (BDSG): Sample Generation on the Boundary}
\name{Nikolaos Dionelis, Mehrdad Yaghoobi, Sotirios A. Tsaftaris}
\address{The University of Edinburgh, Edinburgh, UK}
\begin{document}
\maketitle
\begin{abstract}
Generative models, such as Generative Adversarial Networks (GANs), have been used for unsupervised anomaly detection. While performance keeps improving, several limitations exist particularly attributed to difficulties at capturing multimodal supports and to the ability to approximate the underlying distribution closer to the tails, i.e. the boundary of the distribution's support. This paper proposes an approach that attempts to alleviate such shortcomings. We propose an invertible-residual-network-based model, the Boundary of Distribution Support Generator (BDSG). GANs generally do not guarantee the existence of a probability distribution and here, we use the recently developed Invertible Residual Network (IResNet) and Residual Flow (ResFlow), for density estimation. These models have not yet been used for anomaly detection. We leverage IResNet and ResFlow for Out-of-Distribution (OoD) sample detection and for sample generation on the boundary using a compound loss function that forces the samples to lie on the boundary. The BDSG addresses non-convex support, disjoint components, and multimodal distributions. Results on synthetic data and data from multimodal distributions, such as MNIST and CIFAR-10, demonstrate competitive performance compared to methods from the literature.
\end{abstract}


\section{Introduction}
\label{sec:intro}
Anomaly detection is the identification of
samples different from typical data \cite{DeepGenerativeModels, DeepGenerativeModels2}. When anomalies are not known in advance, unsupervised learning with generative
models is used. 
The aim is to learn a model of ``normality'' with anomalies being  detected
as deviations from this model \cite{WAIC, Deecke}. Important
goals are reducing misdetections and false alarms, estimating the
support of the ``normal'' data distribution, detecting anomalies
close to the support boundary, generating within-distribution
and Out-of-Distribution (OoD) data, and providing decision boundaries
for inference of within and OoD.

Existing approaches to anomaly detection use probability, reconstruction \cite{GANomaly,SkipGANomaly}, and domain based models. GANs are trained to generate samples and fit the ``normal'' data distribution \cite{SGAN, MarioLucic}. During inference, an anomaly score of a queried test sample, $\textbf{x}^*$, is computed by evaluating the probability of obtaining
$\textbf{x}^*$ with the generator \cite{StanfordCourse-1}. Such
models belong to the probability-based methods (e.g. AnoGAN)
\cite{AD1, BiGAN2}. However, these models do not directly address the major problems of multimodal support and the ability to generate on the tails/boundaries. Recent approaches have tried to improve performance and alleviate shortcomings (e.g. MinLGAN and FenceGAN) \cite{MinLGAN, FenceGAN}. At present, generative models based on invertible residual networks, such as \cite{MinLGAN-1,MinLGAN-2}, are lacking for unsupervised anomaly
detection \cite{MemorizingNormality, MultipleHypotheses}. Anomaly detection techniques show discernible limitations for detecting anomalies near the support of multimodal distributions \cite{DeepAEs-1, BiGAN3}.

This work aims at addressing these limitations. Our
aim is to detect abnormalities and generate samples on the boundary of the underlying multimodal distribution of the ``normal data''. We train
invertible models \cite{MinLGAN-1}
to estimate the density of typical samples and propose a loss function
for the boundary generator. We pay particular attention to anomalies close to the boundary of the data distribution and to anomalies near high-probability normal samples. We focus on the ability to model multimodal distributions
with non-convex support and disjoint components. Our model is denoted by Boundary of Distribution
Support Generator (BDSG). It achieves competitive performance on synthetic and typically used benchmark data. In summary, our contributions are: (a) Training invertible generative models and evaluating
the use of inference for anomaly detection, and (b) Sample generation on the tails.

\section{Related Work: Boundary Generation}
\label{sec:format}
The GAN discriminator estimates the distance between the target and model distributions, while the generator learns the mapping from the latent space, $\textbf{z}$, to the data
space, $\textbf{x}$. The GAN optimization is $\text{argmin}_{\pmb{\theta}_g} \text{dist}( p_{\textbf{x}}(\textbf{x}), p_g(\textbf{x}) )$, where the distance metric is given by $\text{argmax}_{\pmb{\theta}_d} \, f( D, p_{\textbf{x}}(\textbf{x}), p_g(\textbf{x}) )$, e.g. $\text{dist}(\textbf{x}, \textbf{y}) = ||\textbf{x} - \textbf{y} ||_{\infty} = \text{max}_i | x_i - y_i |$. The GAN loss is\begin{align}
\begin{split}
\text{argmin}_{\pmb{\theta}_g} & \text{argmax}_{\pmb{\theta}_d} \E_{\textbf{x}} [\text{log}(D(\textbf{x}; \pmb{\theta}_d))] \\
& + \E_{\textbf{z}} [\text{log}(1 - D(G(\textbf{z}; \pmb{\theta}_g); \pmb{\theta}_d))]\text{,}
\end{split}
\end{align}

{\setlength{\parindent}{0cm}where $\textbf{z} \sim p_{\textbf{z}}(\textbf{z})$, $\textbf{x} \sim p_{\textbf{x}}(\textbf{x})$, \text{and} $G(\textbf{z}) \sim p_g(\textbf{x})$.} To perform anomaly detection, we need to change (1) and create a discriminator that can distinguish normal from abnormal. Yet, this implies the ability to have learned all underlying modes and have covered the full support of the distribution from limited data. Unfortunately, GANs tend to learn the mass of the underlying multimodal distribution well, focusing less towards the low probability regions, i.e. the tails, and have discernible problems with mode collapse \cite{FictitiousGAN, BiGAN3}.

MinLGAN uses minimum likelihood regularization to generate data on
the tail of the normal data distribution \cite{MinLGAN}. FenceGAN performs both sample
generation on the boundary and anomaly detection using the generator
and discriminator, respectively \cite{FenceGAN}. The generator loss is reinforced with bespoke losses to help model the boundary and the output of the discriminator is used as an anomaly
threshold. However, FenceGAN does not succeed to form multimodal supports and to detect anomalies near discontinuous boundaries.
\begin{figure}[t]
\centering\includegraphics[clip,width=1\columnwidth]{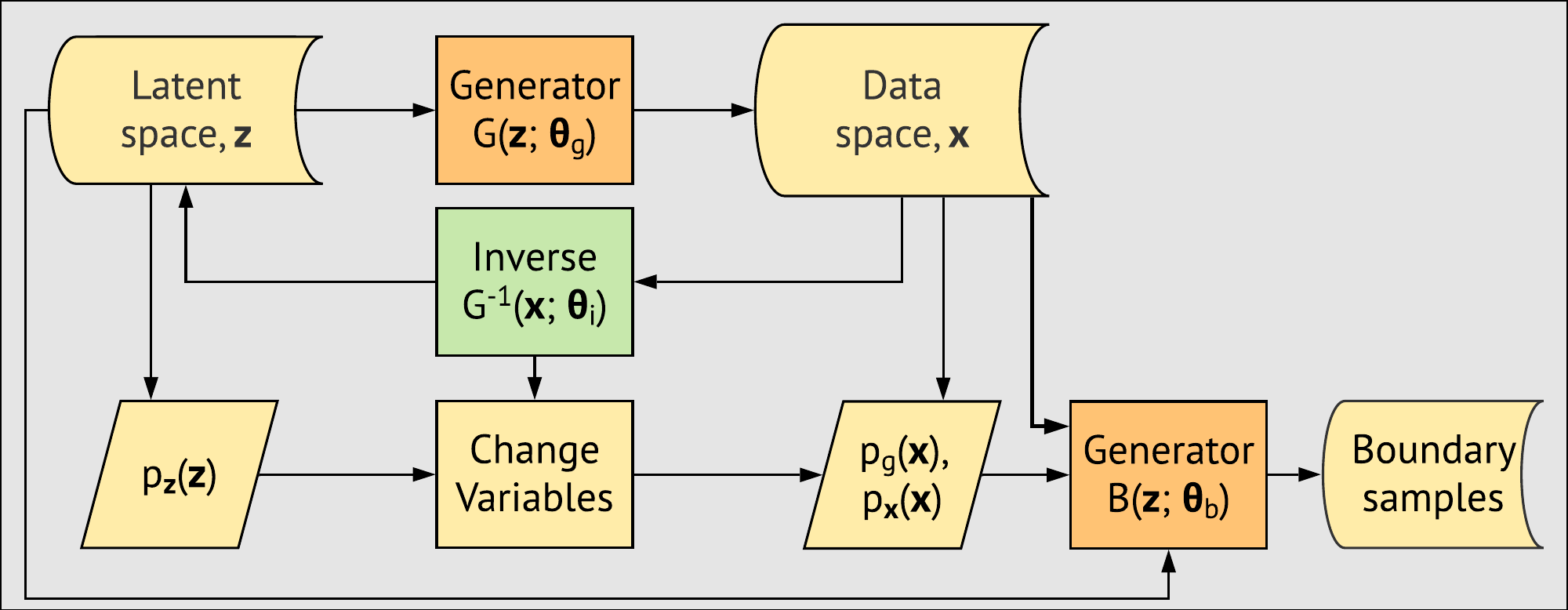}\caption{Flowchart of BDSG to learn the mapping $B: \textbf{z} \rightarrow \textbf{x}$.}
\label{fig:EDCs-1-2-1-2-1-1-1-2-1-1-1-2-1-1-1-1-1-1-1-1-2-5-001-1-3-1-1-1-1} 
\end{figure}

\section{The Proposed BDSG Model}
\label{sec:pagestyle}
We propose the BDSG to detect strong anomalies which are near the
boundary of the normal data distribution. The BDSG flowchart is shown
in Fig.~\ref{fig:EDCs-1-2-1-2-1-1-1-2-1-1-1-2-1-1-1-1-1-1-1-1-2-5-001-1-3-1-1-1-1}. The premise of our approach is to use two generators: $G$ models data of the distribution and $B$ models data that lie close to the support boundary of the distribution. Specifically, we first train an invertible generator, $G(\textbf{z})$, in the form of IResNet \cite{MinLGAN-1}  and
ResFlow \cite{GLoW-1}, $G(\textbf{z}) \sim p_g(\textbf{x})$. $\textbf{z}$ follows a standard Gaussian distribution, $\textbf{z} \sim N(\textbf{0}, \textbf{I})$,
and the mapping from the latent space, $\textbf{z} \in \mathbb{R}^d$,
to the data space, $\textbf{x} \in \mathbb{R}^d$, is given by $G(\textbf{z})$.
The inverse is given by $G^{-1}(\textbf{x})$. The second step is
to train a generator, $B(\textbf{z})$, to perform sample generation
on the support boundary of the data distribution, learning the
mapping $B: \textbf{z} \rightarrow \textbf{x}$.

We now formulate the BDSG loss function. The first term, $L_0(\pmb{\theta}_b, \textbf{z})$, guides to find the boundary, while the second term, $L_1(\pmb{\theta}_b, \textbf{z})$, penalizes deviations from the ``normal class'' using the distance from a point to a set. The third term, $L_2(\pmb{\theta}_b, \textbf{z})$, is for the scattering of the $B(\textbf{z})$ samples in the $\textbf{x}$ space.
$L_2(\pmb{\theta}_b, \textbf{z})$
is for dispersion and diversity and is the ratio of distances in the $\textbf{z}$ and $\textbf{x}$ spaces.
With $L_2(\pmb{\theta}_b, \textbf{z})$, BDSG addresses the mode collapse problem. The loss function for $B(\textbf{z}; \pmb{\theta}_b)$ is\begin{align}
& \text{argmin}_{\pmb{\theta}_b} \, L(\pmb{\theta}_b, \textbf{z}, \textbf{x}, G, \lambda_1, \lambda_2)\text{,}
\end{align}

{\setlength{\parindent}{0cm}where the loss, $L(\pmb{\theta}_b, \textbf{z}, \textbf{x}, G, \lambda_1, \lambda_2)$,
is given by}\begin{align}
& L_0(\pmb{\theta}_b, \textbf{z}, G) + \lambda_1 L_1(\pmb{\theta}_b, \textbf{z}, \textbf{x}) + \lambda_2 L_2(\pmb{\theta}_b, \textbf{z})
\end{align}\begin{align}
\begin{split}
& = \frac{1}{N} \sum_{i=1}^N \left[ p_g( B(\textbf{z}_i; \pmb{\theta}_b) ) + \lambda_1 \min_{j=1}^M || B(\textbf{z}_i; \pmb{\theta}_b) - \textbf{x}_j ||_2 \right. \\
& \left. + \lambda_2 \frac{1}{N-1} \sum_{j=1, \, j \neq i}^N \frac{ || \textbf{z}_i - \textbf{z}_j||_2 }{ || B(\textbf{z}_i; \pmb{\theta}_b) - B(\textbf{z}_j; \pmb{\theta}_b) ||_2 } \right]\text{,}
\end{split}
\label{eq:eqeq31}
\end{align}
\label{eq:eqeq21}

{\setlength{\parindent}{0cm}where $\lambda_1$ and $\lambda_2$ are
hyper-parameters of the BDSG. In (3) and (4), the first term,
$L_0(\pmb{\theta}_b, \textbf{z}, G)$, is given by}\begin{align}
& \frac{1}{N} \sum_{i=1}^N \left[ p_{\textbf{z}}(G^{-1}(B(\textbf{z}_i; \pmb{\theta}_b))) \left | \text{det} \, \textbf{J}_{G}(B(\textbf{z}_i; \pmb{\theta}_b)) \right |^{-1} \right] \\
\begin{split}
& = \frac{1}{N} \sum_{i=1}^N \left[ \exp ( \log (p_{\textbf{z}}(G^{-1}(B(\textbf{z}_i; \pmb{\theta}_b)))) \right. \\
& \left. \ \ \ - \log ( \left | \text{det} \, \textbf{J}_{G}(B(\textbf{z}_i; \pmb{\theta}_b)) \right | ) ) \right]\text{,}
\end{split}
\end{align}

{\setlength{\parindent}{0cm}where $\log (p_{\textbf{z}}(G^{-1}(B(\textbf{z}_i))))$
and $\log ( \left | \text{det} \, \textbf{J}_{G}(B(\textbf{z}_i)) \right | )$
are estimated by an invertible model. The $\pmb{\theta}_b$
parameters are obtained by running Gradient Descent on $L(\pmb{\theta}_b, \textbf{z}, \textbf{x})$,
which can decrease to zero and is written in terms of the sample size,
$M$, and the batch size, $N \leq M$. In the loss in \eqref{eq:eqeq31},
the effective dimensionality of $\, \textbf{z} \sim \text{N}(\textbf{0}, \textbf{I}) \,$
is lower than that of $\, \textbf{x}$.}

\subsection{BDSG Benefits in Sampling Complexity, Anomaly Detection, and Generation of Strong Anomalies}
\label{sec:typestyle}
\textbf{The Sampling Complexity Problem:} To perform anomaly detection, FenceGAN estimates $p(\textbf{x} | D(\textbf{x}) < \delta)$. This is difficult due to the rarity problem since at least $\delta^{-2}$
points are needed on the tail of the distribution. Sampling from a
distribution could fail to have even a single point in low probability
regions \cite{ADwithGANs-1-1, ADwithGANs-1}. However, the FenceGAN loss does not succeed to generate a discrete boundary around multimodal distributions separately because it is based on the parallel simultaneous estimation of the density and of sample generation on the boundary. 
In contrast, the proposed BDSG obviates the rarity problem achieving better sampling complexity.

\textbf{Anomaly Detection:} During inference, a test sample, $\textbf{x}^*$, is anomalous if $p_g(\textbf{x}^*)=0$ and normal otherwise. In practice, a threshold, $\epsilon$, is used instead of $0$. The first term of the loss in \eqref{eq:eqeq31} discriminates between normal and abnormal data.

\textbf{Generating Strong Anomalies:} The BDSG can generate samples lying on the tail of the data distribution, i.e. strong anomalies. First, the boundary generator, $B(\textbf{z})$, generates $Q >> N$ samples. Then, the probability of each of these boundary samples is computed and $L_0(\pmb{\theta}_b, \textbf{z}, G)$
in \eqref{eq:eqeq31}, and if $p_g(B(\textbf{z})) < \epsilon$, then
$B(\textbf{z})$ is a strong anomalous sample.

\section{Evaluation of the BDSG}
\label{sec:majhead}
We evaluate BDSG on synthetic and image data considering several criteria that measure its ability to approximate the boundary and detect anomalies. We evaluate
the BDSG for anomaly detection using the Area Under the Receiver Operating
Characteristics Curve (AUROC) and the Area Under the Precision-Recall
Curve (AUPRC). Using the leave-one-out methodology, we compare the
BDSG with the state-of-the-art models of GANomaly, AnoGAN, MinLGAN, and FenceGAN on MNIST, CIFAR-10, and other datasets for OoD.
\begin{figure}[t]
\begin{minipage}[t]{0.48\columnwidth}%
\centering \includegraphics[bb=10bp 0bp 420bp 320bp,clip,width=1\columnwidth]{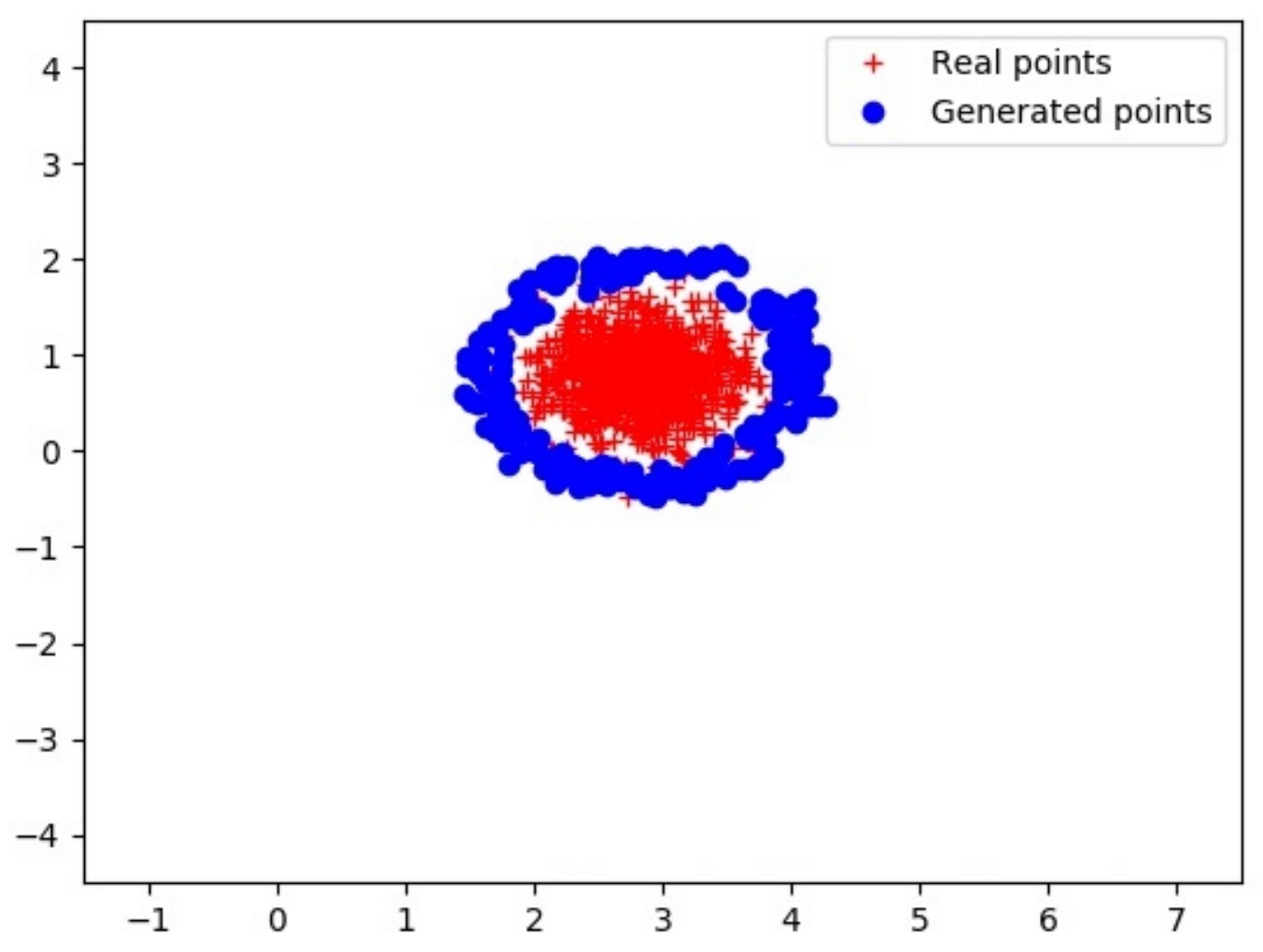}

\vspace{-2pt}

{\footnotesize (a)}%
\end{minipage}\hfill{}%
\begin{minipage}[t]{0.48\columnwidth}%
\centering \includegraphics[bb=10bp 0bp 578bp 430bp,clip,width=1\columnwidth]{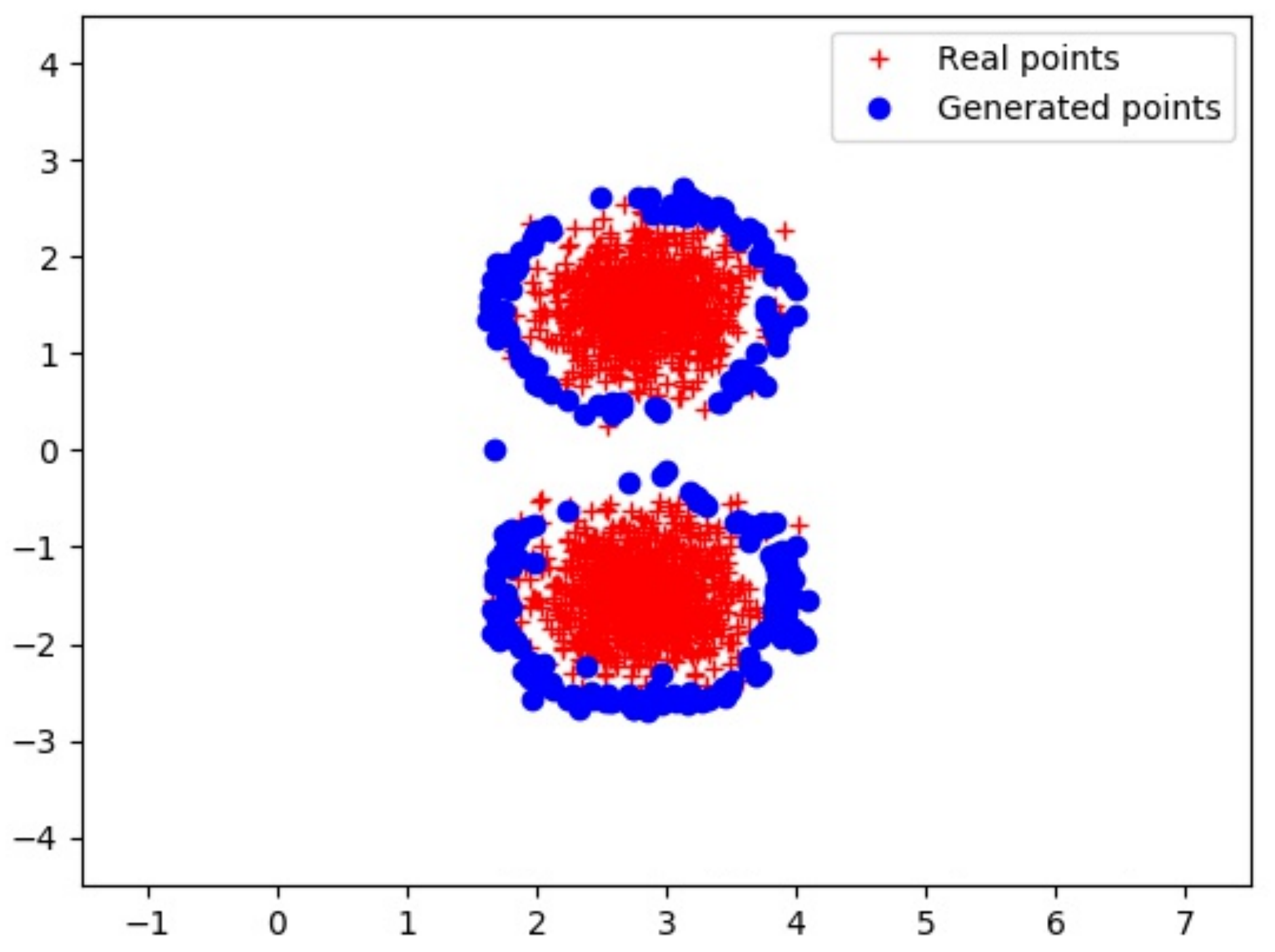}

\vspace{-2pt}

{\footnotesize (b)}%
\end{minipage}


\begin{minipage}[t]{0.48\columnwidth}%
\centering \includegraphics[bb=10bp 0bp 578bp 430bp,clip,width=1\columnwidth]{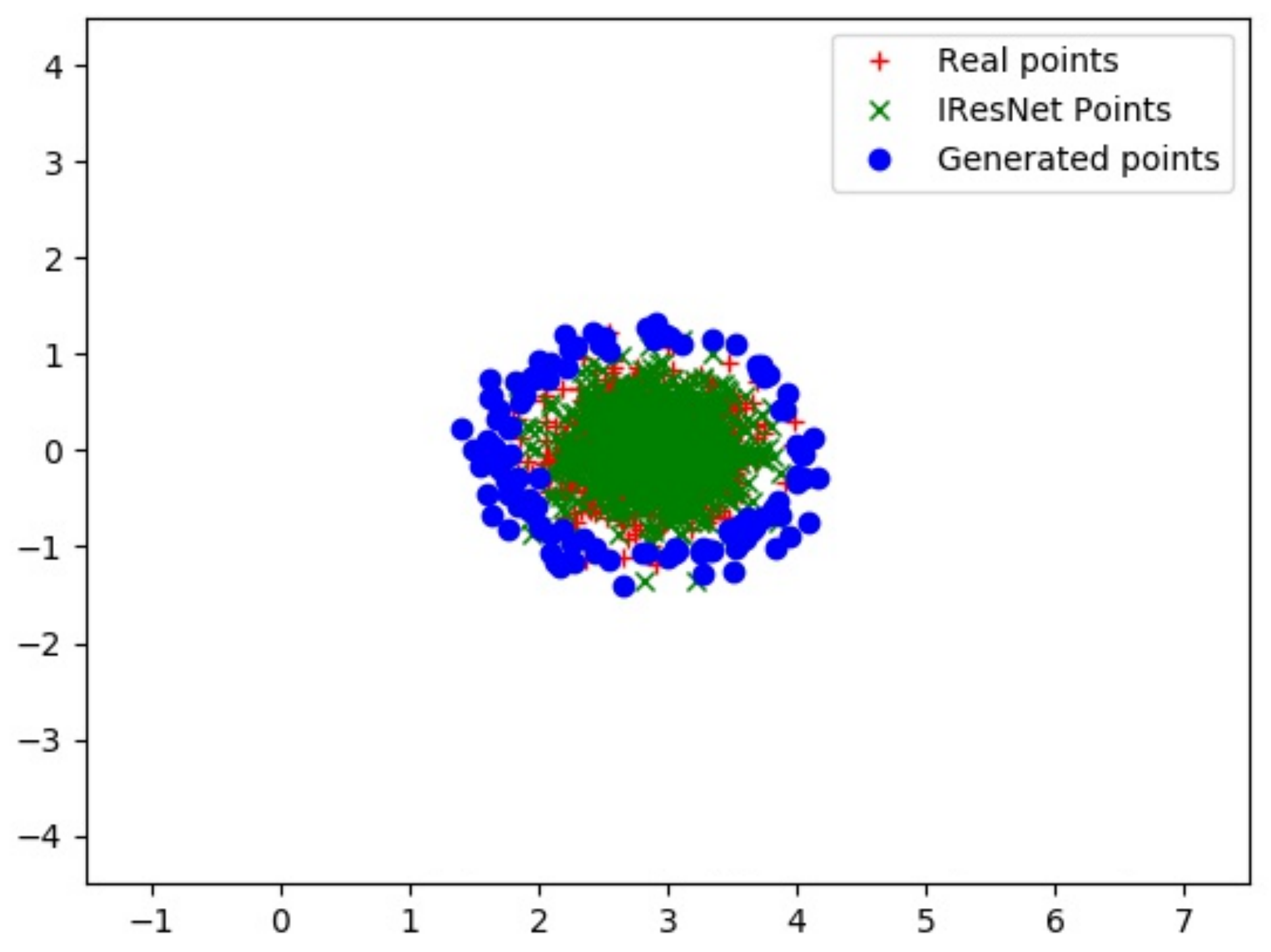}

\vspace{-2pt}

{\footnotesize (c)}%
\end{minipage}\hfill{}%
\begin{minipage}[t]{0.48\columnwidth}%
\centering \includegraphics[bb=10bp 0bp 420bp 320bp,clip,width=1\columnwidth]{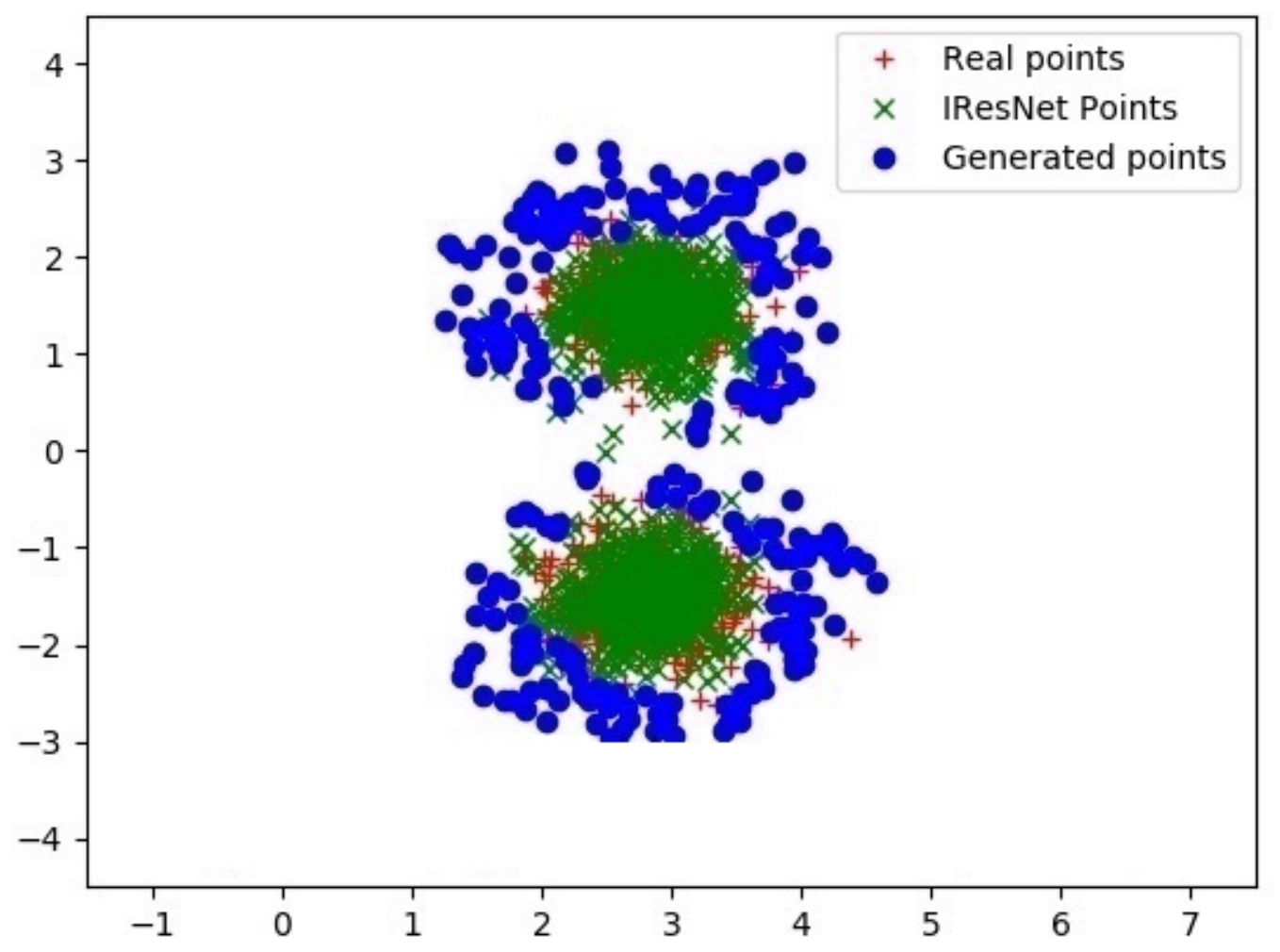}

\vspace{-2pt}

{\footnotesize (d)}%
\end{minipage}

\caption{(a,b) CFS BDSG for uni and multimodal distributions. (c,d) IResNet
BDSG. The red points are samples, the green points IResNet samples, and the blue points BDSG samples.}

\label{fig:EDCs-1-2-1-2-1-1-1-2-1-1-1-2-1-1-1-1-1-1-1-1-2-5-001-1-3-1-1-1-1-1-1}

\label{fig:EDCs-1-2-1-2-1-1-1-2-1-1-1-2-1-1-1-1-1-1-1-1-2-5-001-1-3-1-1-1-1-1-1-1}

\label{fig:EDCs-1-2-1-2-1-1-1-2-1-1-1-2-1-1-1-1-1-1-1-1-2-5-001-1-3-1-1-1-1-1-1-1-2}

\label{fig:EDCs-1-2-1-2-1-1-1-2-1-1-1-2-1-1-1-1-1-1-1-1-2-5-001-1-3-1-1-1-1-1-1-1-1}

\label{fig:EDCs-1-2-1-2-1-1-1-2-1-1-1-2-1-1-1-1-1-1-1-1-2-5-001-1-3-1-1-1-1-1-1-1-1-1}

\label{fig:EDCs-1-2-1-2-1-1-1-2-1-1-1-2-1-1-1-1-1-1-1-1-2-5-001-1-3-1-1-1-1-1-1-1-1-1-1}
\end{figure}

\textbf{Setup:} \textit{Synthetic data:}  We test BDSG
using two experimental setups using the multivariate Gaussian distribution, where we know the closed-form of the underlying probability density function. The first setup uses a closed-form solution (CFS) evaluation of $L_0(\pmb{\theta}_b, \textbf{z}, G)$ 
model distribution, in lieu of $p_g(\textbf{x})$. The second setup uses $p_g(\textbf{x})$ from IResNet \cite{MinLGAN-1}.

\textit{Benchmark data:} We also evaluate the BDSG on MNIST by first training an invertible generator, ResFlow, for density estimation. We then train the BDSG using a convolutional neural network (CNN), applying \eqref{eq:eqeq31}. Then, we evaluate the performance of the BDSG on CIFAR-10. Further, we evaluate the performance of the BDSG trained on MNIST
and CIFAR-10 and tested on OoD data using the algorithm convergence criteria of the proposed loss and its second term, $L_1$ \cite{AdamOptimizer-2-1}.

\textit{Models:} We use a fully-connected $B(\textbf{z})$ model for synthetic data and CNN and batch normalization for images.

\subsection{B(\textbf{z}) Model Architecture for Synthetic Data}
\textbf{CFS BDSG Model:} Based on sensitivity analyses, we use dense fully-connected
layers for $B(\textbf{z})$, $M=1024$, $N=256$, $\lambda_1=0.3$, and $\lambda_2=0.025$.
The sample size, $M$, affects the BDSG performance. The batch size,
$N$, affects the convergence speed and can lead to a thinner boundary.
Figure~\ref{fig:EDCs-1-2-1-2-1-1-1-2-1-1-1-2-1-1-1-1-1-1-1-1-2-5-001-1-3-1-1-1-1-1-1}(a)
shows the boundary formed using the CFS BDSG for a unimode distribution.
The red points are from the normal data distribution; the blue $B(\textbf{z})$
points are on the estimated boundary. The 2-8-8-2 model for $B(\textbf{z})$
achieves a low loss function value and converges the samples to the boundary. For a bimodal distribution in Fig.~\ref{fig:EDCs-1-2-1-2-1-1-1-2-1-1-1-2-1-1-1-1-1-1-1-1-2-5-001-1-3-1-1-1-1-1-1-1}(b),
a 2-8-8-8-2 network leads to low loss values and accurate boundary formation. The average probability of the $B(\textbf{z})$ points, which are on the boundary, is $L_0(\pmb{\theta}_b, \textbf{z}, G) = 0.007$ in (3). We obtain descending loss values, successfully converging $B(\textbf{z})$ to the boundary.

\textbf{IResNet-Based BDSG:} To show that BDSG yields competitive performance on synthetic data from multimodal distributions, we also perform a second experiment. We train our chosen invertible model, IResNet, and use the estimated density to create the boundary.
If $p_g(\textbf{x})$ is estimated correctly, then BDSG
estimates the boundary of $p_{\textbf{x}}(\textbf{x})$. In Fig.~\ref{fig:EDCs-1-2-1-2-1-1-1-2-1-1-1-2-1-1-1-1-1-1-1-1-2-5-001-1-3-1-1-1-1-1-1-1-1}(c),
we use a 2-8-8-8-2 network for $B(\textbf{z})$ for the unimode distribution, $M=1024$,
$N=128$, $\lambda_1 = 0.3$, and $\lambda_2 = 0.025$.

For the bimodal distribution
in Fig.~\ref{fig:EDCs-1-2-1-2-1-1-1-2-1-1-1-2-1-1-1-1-1-1-1-1-2-5-001-1-3-1-1-1-1-1-1-1-1-1}(d),
we use a deeper architecture for $B(\textbf{z})$, $N=256$, and $\lambda_2 = 0.25$.
An ablation study
found that $L_2(\pmb{\theta}_b, \textbf{z})$ in \eqref{eq:eqeq31}
is necessary, and otherwise mode collapse is encountered. In Fig.~2(d), for evaluation, we also use the boundary clustering algorithm given by\begin{align}
& \text{argmin}_{i = 1, \dots, K} \, \text{min}_{ j = 1, \dots, L} \, || B(\textbf{z}) - \textbf{x}_{i,j} ||_2\text{,}
\end{align}

{\setlength{\parindent}{0cm} where $K=2$ clusters from the
bimodal distribution, $L=0.5M$ samples from each mode, and $\textbf{x}_{i,j} \in \mathbb{R}^d$
is the $j$-th sample of mode $i$. Here, $\text{min}_{ j = 1, \dots, L} \, || B(\textbf{z}) - \textbf{x}_{i,j} ||_2$
is negligible, smaller than the distance from a mode/set to a set.}

Figures~2(b) and (d) show that BDSG achieves
successful boundary formation and stable convergence without mode
collapse. BDSG is compared to FenceGAN and FenceGAN
yields incomplete boundary formation between the modes.

\subsection{Binary Classification and Boundary Precision}
We create a grid of equidistant points in the 2D space and associate
each grid point with a probability using the distribution in Fig.~2(d). Using a threshold, $\epsilon$,
to detect anomalies, we evaluate the inference performance
of $L_0$ in \eqref{eq:eqeq31} by
computing binary classification metrics. To examine the influence of the choice of $\epsilon$, we compute precision, recall, F1 score, and accuracy, and these scores are higher than $0.99$ for $\epsilon \geq 0.5\%$. To examine how accurate we estimate the boundary and to compare with IResNet, we define two Boundary Precision (BP) scores. By analogy with precision, BP1 is the percentage of $B(\textbf{z})$-points that satisfy
$p_{\textbf{x}}(B(\textbf{z})) \in [ \gamma, \epsilon ]$. BP2 is defined as the intersection of the grid points with IResNet. BP1 is always higher than BP2: $\text{BP1}=80\%$, $\text{BP2}=68\%$ when $\epsilon = 1\%$.

\begin{figure}[t]
\vspace{-8.25pt}

\centering \includegraphics[clip,width=1\columnwidth]{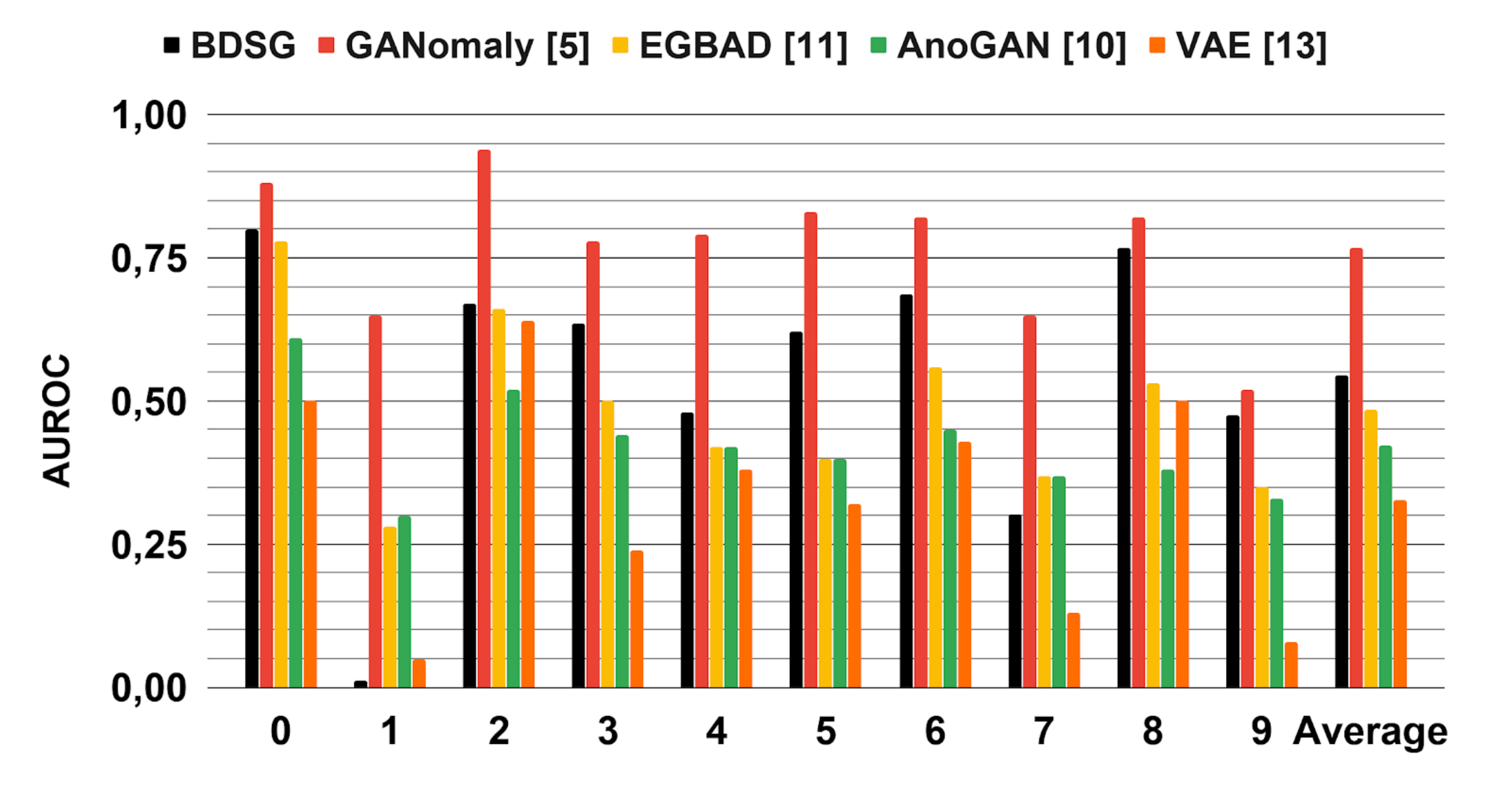}

\vspace{0.25pt}

\centering \includegraphics[clip,width=1\columnwidth]{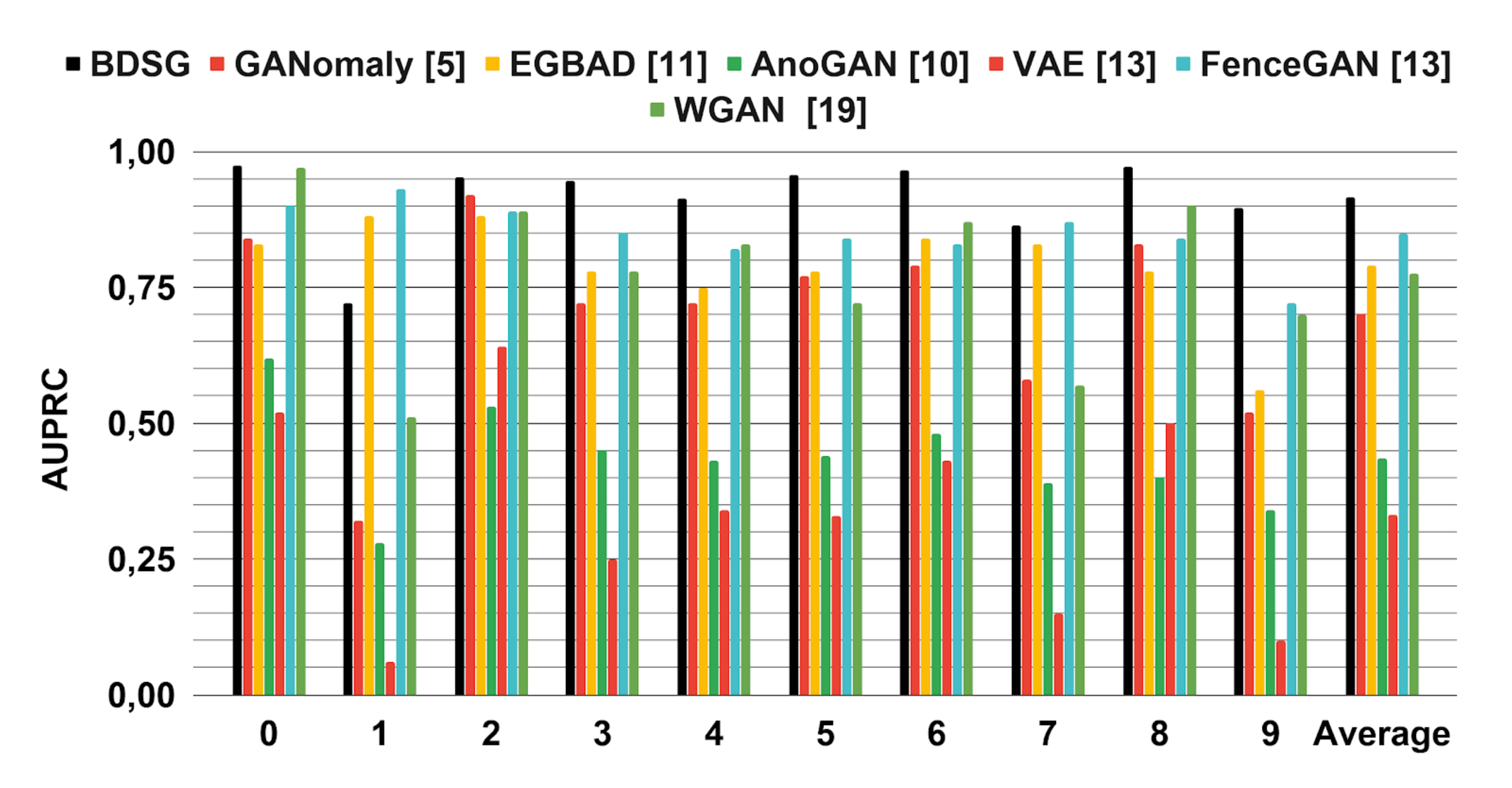}

\vspace{-8pt}

\caption{AUROC and AUPRC evaluation on MNIST data.}

\label{fig:EDCs-1-2-1-2-1-1-1-2-1-1-1-2-1-1-1-1-1-1-1-1-2-5-001-1-3-1-1-1-1-1-1-1-1-2-1}

\label{fig:EDCs-1-2-1-2-1-1-1-2-1-1-1-2-1-1-1-1-1-1-1-1-2-5-001-1-3-1-1-1-1-1-1-1-1-2-1-1}
\end{figure}

\subsection{Evaluation of the BDSG on Image Data}
\textbf{MNIST.} \textit{Setup:} We train ResFlow until convergence on MNIST using the leave-one-out
evaluation where the anomaly class is the leave-out digit and the normal class
is the remaining digits. We then train
the BDSG using a CNN with batch normalization, using \eqref{eq:eqeq31}.
We also examine
different models such as feed-forward and residual. For $M$, we use the entire training set and we also examine different
values for $N$ in \eqref{eq:eqeq31}. After convergence,
the loss is $0.7$, $L_0(\pmb{\theta}_b, \textbf{z}, G) = 0$, and
$L_1(\pmb{\theta}_b, \textbf{z}, \textbf{x}) = 0.8$. This $L_1(\pmb{\theta}_b, \textbf{z}, \textbf{x})$,
which is the distance from a point to a set, is smaller than the minimum set distance of every pair of MNIST digits which is approximately $10$. For evaluation, we compare the proposed BDSG with state-of-the-art models using AUROC and AUPRC as they are commonly used evaluation criteria in the literature \cite{GANomaly}.

\textit{Findings:} Figure~3 shows that BDSG achieves competitive performance compared
to the alternative techniques and on average and for most digits, BDSG outperforms EGBAD, AnoGAN, and VAE in AUROC and GANomaly, EGBAD, AnoGAN, VAE, FenceGAN, and WGAN in AUPRC.

Going beyond the leave-one-out setting, we assess how BDSG performs when other OoD data are used as anomaly samples considering MNIST as normal and Fashion-MNIST and KMNIST as OoD abnormal 
\cite{DeepGenerativeModels}. We report results in Table~1 using algorithm convergence criteria, the
proposed loss and $L_1$.
The loss and $L_1$ are lower for the normal class, digits 1 to 9, than for the anomaly class, digit 0, and the abnormal OoD data indicating that the proposed loss and its first term can be used for anomaly detection with a threshold of $1$.

\textbf{CIFAR-10.} \textit{Setup:} We train ResFlow and IResNet for density estimation on CIFAR-10
\cite{MinLGAN-2}. Next, we train BDSG using a CNN with batch normalization and applying \eqref{eq:eqeq31}.
\begin{table}[t]
\centering

\begin{tabular}{|c|c|c|c|c|c|}
\hline 
{\footnotesize \textbf{MNIST}} & {\footnotesize Loss} & {\footnotesize $L_1$} & {\footnotesize \textbf{CIFAR-10}} & {\footnotesize Loss} & {\footnotesize $L_1$}\tabularnewline
\hline 
\hline 
{\footnotesize Digits 1-9} & {\footnotesize $0.74$} & {\footnotesize $0.93$} & {\footnotesize CIFAR-10} & {\footnotesize $3.16$} & {\footnotesize $8.94$}\tabularnewline
\hline 
{\footnotesize Digit 0} & {\footnotesize $20.36$} & {\footnotesize $66.32$} & {\footnotesize CIFAR-100} & {\footnotesize $7.50$} & {\footnotesize $23.43$}\tabularnewline
\hline 
{\footnotesize Fashion-MNIST} & {\footnotesize $9.92$} & {\footnotesize $31.44$} & {\footnotesize SVHN} & {\footnotesize $7.18$} & {\footnotesize $22.36$}\tabularnewline
\hline 
{\footnotesize KMNIST} & {\footnotesize $9.28$} & {\footnotesize $29.37$} & {\footnotesize STL-10} & {\footnotesize $10.01$} & {\footnotesize $31.75$}\tabularnewline
\hline 
\end{tabular}

\vspace{4pt}

\caption{Evaluation of BDSG comparing normality (MNIST Digits 1-9 and CIFAR-10)
with the abnormal class and the anomaly cases (MNIST Digit 0, Fashion-MNIST,
KMNIST and CIFAR-100, SVHN, STL-10), using $L$ and $L_1$ in \eqref{eq:eqeq31}.}

\label{tab:tabTabTab333331sadfsa3sadfs1sadfsafasdf311sdfsa11}

\label{tab:tabTabTab333331sadfsa3sadfs1sadfsafasdf311sdfsa112}
\end{table}
\begin{figure}[t]

\centering \includegraphics[clip,width=1.0011\columnwidth]{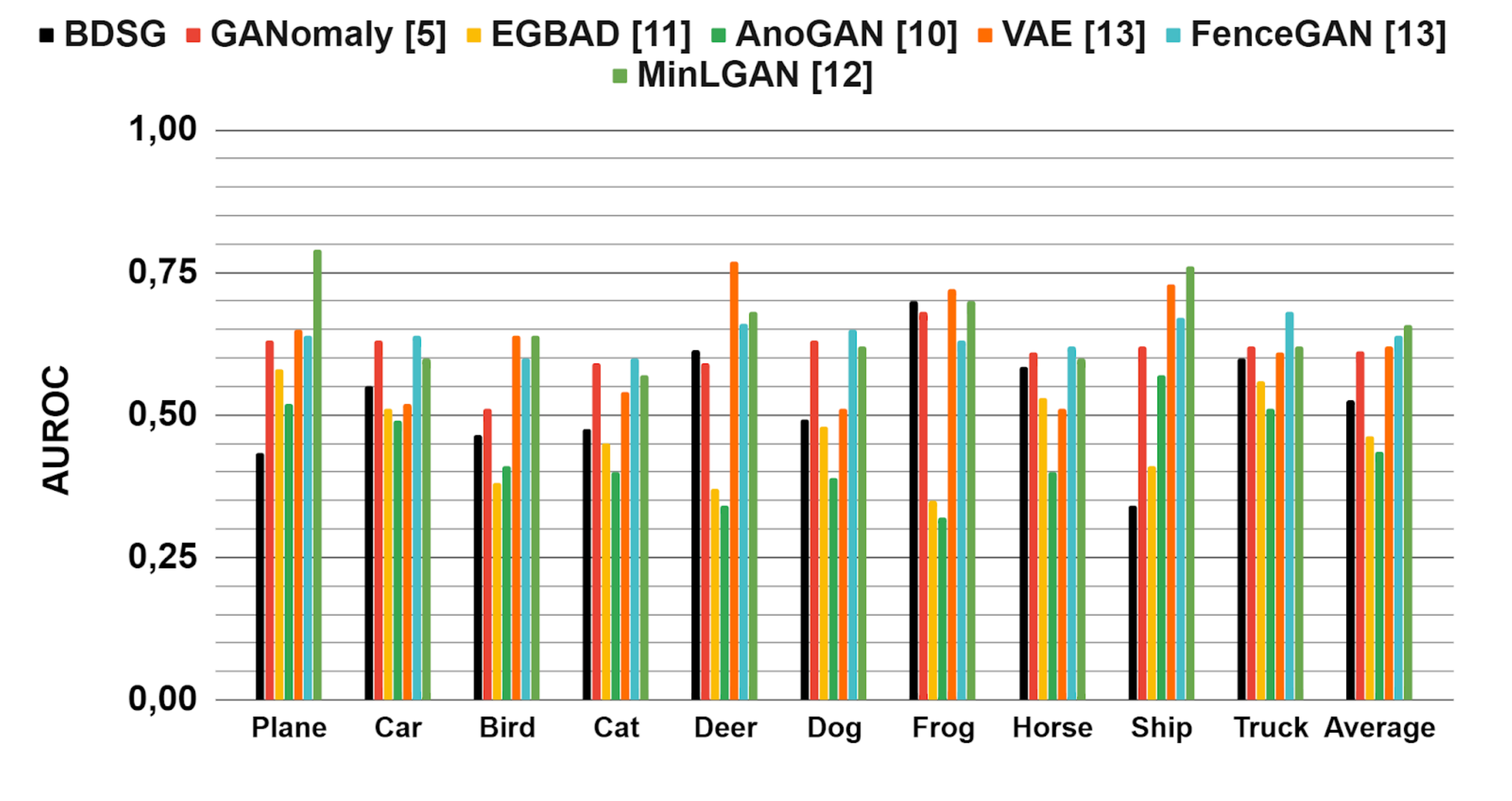}

\vspace{-7.2pt}

\caption{AUROC evaluation of BDSG on CIFAR-10 data.}

\label{fig:EDCs-1-2-1-2-1-1-1-2-1-1-1-2-1-1-1-1-1-1-1-1-2-5-001-1-3-1-1-1-1-1-1-1-1-2-1-3}

\label{fig:EDCs-1-2-1-2-1-1-1-2-1-1-1-2-1-1-1-1-1-1-1-1-2-5-001-1-3-1-1-1-1-1-1-1-1-2-1-1-2}

\label{fig:EDCs-1-2-1-2-1-1-1-2-1-1-1-2-1-1-1-1-1-1-1-1-2-5-001-1-3-1-1-1-1-1-1-1-1-2-1-2}

\label{fig:EDCs-1-2-1-2-1-1-1-2-1-1-1-2-1-1-1-1-1-1-1-1-2-5-001-1-3-1-1-1-1-1-1-1-1-2-1-1-1}
\end{figure}

\textit{Findings:} Figure~4 presents the AUROC for each CIFAR-10 class. On a leave-one-out evaluation, the BDSG outperforms on average EGBAD and AnoGAN. We demonstrate the efficacy of the proposed BDSG model which achieves competitive performance in AUROC compared to EGBAD, AnoGAN, and VAE. Table~1 presents the performance evaluation of
the BDSG to detect abnormal OoD data from CIFAR-100, SVHN, and STL-10
using the algorithm criteria of the loss and $L_1(\pmb{\theta}_b, \textbf{z}, \textbf{x})$.
Both $L$ and $L_1$ in \eqref{eq:eqeq31} are high
for the anomaly cases deviating from normality, indicating that an anomaly detection threshold can be imposed on either the proposed cost or its second term, e.g. $4$ on $L$ and $9$ on $L_1$.

\section{Conclusion}
For anomaly detection, the accurate determination of the support
boundary is critical and in this paper, we present the BDSG which uses
the loss in \eqref{eq:eqeq31} and leverages reversibility to compute
the probability at any point in $\textbf{x}$. It addresses the rarity
problem and the detection of strong anomalies, and maps from $\textbf{z}$
to $\textbf{x}$ concentrating the images of $\textbf{z}$ on the
boundary. Using invertible models has advantages in improving
the anomaly detection methodology by allowing to devise a generator for creating boundary samples. The BDSG performs sample generation
on the boundary, addresses mode collapse, and achieves competitive performance
on synthetic data from multimodal distributions and on MNIST and
CIFAR-10.

\section{Acknowledgment}
This work was supported by the UK EPSRC Grant Number EP/S000631/1 and the UK MOD
UDRC in Signal Processing.


\bibliographystyle{IEEEbib}
\bibliography{strings,refs}

\end{document}